# Development of a conversing and body temperature scanning autonomously navigating robot to help screen for COVID-19

Ryan H. Kim[1]

[1] Choate Rosemary Hall, Wallingford, Connecticut, USA

**Abstract:** Throughout the COVID-19 pandemic, the most common symptom displayed by patients has been a fever, leading to the use of temperature scanning as a preemptive measure to detect potential carriers of the virus. Human employees with handheld thermometers have been used to fulfill this task, however this puts them at risk as they cannot be physically distanced and the sequential nature of this method leads to great inconveniences and inefficiency. The proposed solution is an autonomously navigating robot capable of conversing and scanning people's temperature to detect fevers and help screen for COVID-19. To satisfy this objective, the robot must be able to (1) navigate autonomously, (2) detect and track people, and (3) get individuals' temperature reading and converse with them if it exceeds 38ºC. An autonomously navigating mobile robot is used with a manipulator controlled using a face tracking algorithm, and an end effector consisting of a thermal camera, smartphone, and chatbot. The goal of this project is to develop a functioning solution that performs the above tasks. In addition, technical challenges encountered, and their engineering solutions will be presented, and recommendations will be made for enhancements that could be incorporated when approaching commercialization.

## 1. INTRODUCTION

Throughout the COVID-19 pandemic, the most common symptom displayed by patients has been a fever, leading many governments and organizations around the world to implement temperature scanning as a preemptive measure of detecting potential carriers of the virus. This procedure was conducted by having human employees manually measure each person's temperature using a handheld thermometer. Additionally, some countries and organizations have begun using temperature scanning kiosks, but both of the above solutions have many critical issues that must be resolved:

- Employee safety. With human employees, they are put at risk as they cannot be socially distanced when taking forehead temperatures.
- People need to manually align their face to the immobile infrared camera of a temperature scanning kiosk. This is both inconvenient and can be time consuming when misalignment leads to either incorrect or no detection.
- The scanner lacks mobility. Entry points have to be limited in order to systematically scan everybody, creating bottlenecks and long waiting times. These compound the issues of inconvenience and inefficiency.
- Immobility means multiple kiosks are required to cover large areas.
- The location of the detector can lead to false negative readings**.** During the winter, the colder outside temperature temporarily lowers the face's temperature, allowing people with a fever to pass as normal.

The goal of this project is to develop an autonomously navigating robot that can converse and scan for people's temperature. Compared to the previously used methods the proposed robotic system is:

- More convenient and timesaving. Autonomous navigation eliminates the need for long lines and a manipulator allows for the system to adjust the temperature scanner without needing the user to move and align their face.
- More accurate temperature readings. Instead of checking someone's temperature once at the gate, repetitive scanning can occur at room temperature without the outside cold artificially lowering body temperature.
- Safety of employees. No people are involved throughout the temperature screening process, protecting them from hazards and violence.

## 2. ROBOTIC SYSTEM ARCHITECTURE

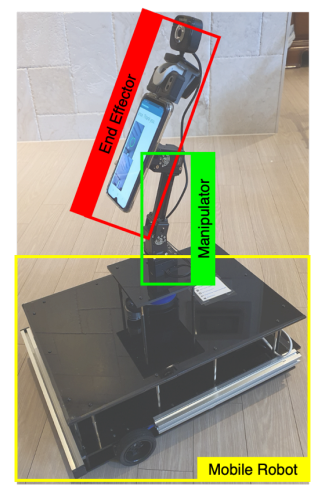

Fig. 1. Full robotic system architecture with the end effector sub-module in red, the manipulator sub-module in green, and the mobile robot sub-module in yellow.

To satisfy the aforementioned goals, the robotic system must be able to:

- Navigate and path plan autonomously

- Accurately find and follow human faces
- Extract temperature values
- Converse with users about symptoms of COVID-19

These tasks can be accomplished using three sub-modules: the end effector, the manipulator, and the mobile robot. The end effector sub-module uses a custom app and a thermal camera with a programmable software development kit (SDK) to extract people's temperature readings [3]. If the temperature exceeds the CDC-defined fever threshold of 38°C, a natural language understanding (NLU) artificial intelligence (AI) chatbot is activated that converses with users about symptoms of COVID-19. The manipulator uses an object-recognition algorithm to detect people and a custom inverse kinematics algorithm is used to orient the manipulator and end effector to align with a detected person's forehead. The mobile robot is used for autonomous navigation and path planning. The mobile robot can detect and avoid obstacles and stops when a person is detected. The Android operating system is used to control the end effector and the middleware called Robot Operating System (ROS) [6] is used to control the manipulator and mobile base. The versions used are Android 10 and ROS 1 Melodic Morenia.

## 2.1 End Effector

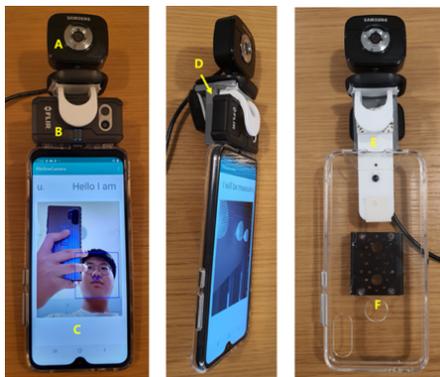

Fig. 2. End effector sub-module with (A) Samsung USB camera, (B) FLIR ONE Pro thermal camera, (C) Samsung Galaxy M20 with custom app, (D) 3D printed fixture, (E) Coat hanger used to secure thermal camera, (F) Phone case used to secure smartphone.

The end effector, attached to the manipulator, is used to scan people's temperature and converse with them if they have a fever. The sub-module consists of four key components:
- Thermal camera
- USB camera
- Android smartphone with custom app
- Natural Language Understanding (NLU) Chatbot

A thermal camera that was affordable and provided an Android-compatible SDK was needed, so the FLIR ONE Pro was selected [3]. The camera is attached to the smartphone using a USB-C port and it is able to provide accurate temperature values when given coordinates.

The USB camera attached to the end effector is used to provide real-time images for the object-recognition algorithm used by the manipulator to detect people.

The custom Android app is developed from the FLIR SDK and consists of three key parts:
- User Interface (UI)
- Interaction with thermal camera
- Interaction with chatbot

The UI is used to present temperature data to users. Extending the view class, a custom camera view is placed in the center of the screen to show the view from the front facing camera. A marquee text lies above this camera view and cycles the words: *"Hello I am a mobile temperature scanner. I will be measuring your temperature and checking for a fever. Please wear your mask and socially distance. Thank you."* This is verbally repeated every ten seconds by using the timer class and a text-to-speech (TTS) engine. The purpose of this marquee text and TTS engine is to inform users of the intent of the robot, to discourage any suspicion or mistrust that may lead to violence.

The Google Mobile Vision API provides face detection and the FaceGraphic.java program receives the data and draws a bounding box around each face. The FLIR thermal camera's ThermalImage thermalImage provides its own image with temperature values, upon which the coordinates and bounding box are mapped. However, because of the smaller 160x120 resolution of the FLIR camera compared to the larger 480x640 pixel resolution of the Samsung USB camera, the dimensions of the bounding box had to be constricted to match the resolution of the FLIR camera. Temperature values from the provided coordinate and eight points around it are collated and the maximum temperature among those nine points on the face is returned.

When a temperature exceeding 38°C is detected, the screen is changed and a natural language understanding chatbot is activated. The chatbot is developed using Google's Dialogflow framework and converses with users about vaccinations and potential symptoms of COVID-19. When provided a series of pre-defined intents and 10 ~ 15 sample phrases, the framework trains itself to accurately extract the meaning behind each of the phrases and matches them to the intent. Machine learning and natural language understanding are used to parse out the meaning of user's inputs. If this aligns with a pre-defined intent, then the chatbot responds with the pre-defined response. If not, the system responds with phrases like *"I'm sorry I didn't understand what you said"* and prompts the user repeat their command.

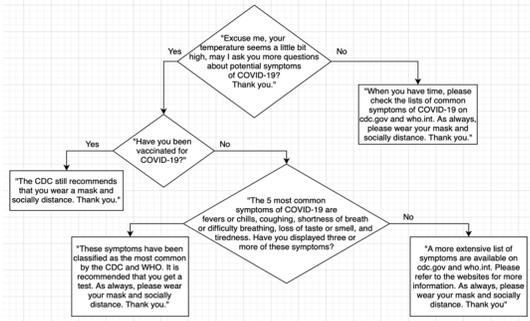

Fig. 3. Chatbot conversation flow

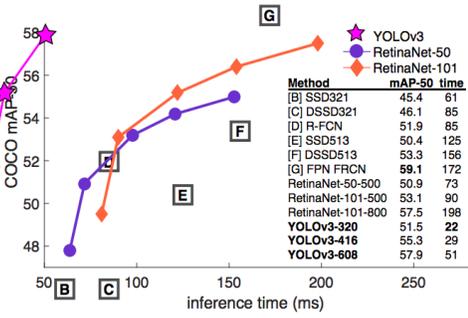

Fig. 5. Comparison of YOLOv3 with RetinaNet-50 and RetinaNet-101 on speed (ms) versus accuracy (mAP-50) on COCO test-dev [5]

The chatbot is able to receive input by either the on-screen keyboard or through a speech-to-text (STT) engine. The chatbot responds in both text, which is displayed on the screen, and verbally using a TTS engine. The conversation flow of the chatbot is as shown in Fig 3.

For the physical build of the sub-module, the smartphone was inverted to reduce the distance between the FLIR ONE Pro thermal camera and the Samsung USB camera to minimize parallax. The Android phone is held in place by a phone case which is screwed onto the manipulator. Above the device, a plastic coat hanger is used to secure the thermal camera in place. Because there were no objects of the right shape, the computer aided design (CAD) software Tinkercad was used to design and 3D print an attachment upon which for the USB camera was connected.

## 2.2 Manipulator

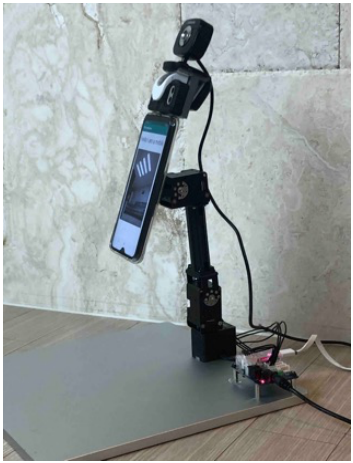

Fig. 4. Manipulator sub-module with end effector

The manipulator is used to detect and follow people to orient the end effector to align with their forehead.

To detect people, a USB camera attached to the end effector was used to send real-time images to the computer onboard the mobile robot. For tracking individuals, a detection algorithm that is both fast and accurate was needed. The You Only Look Once (YOLO) algorithm v3 was selected as it is able to detect images with accuracy comparable to other convolutional neural networks (CNNs) like RetinaNet but at much faster speeds (as shown in Fig. 5).

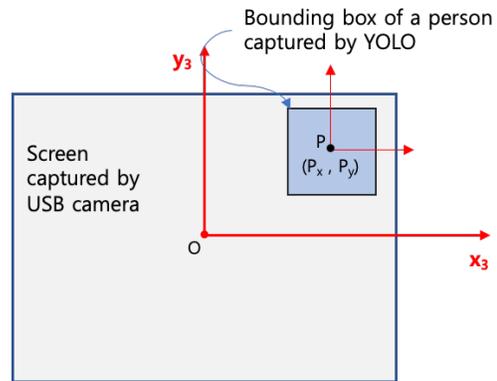

Fig. 6. Aligning the end effector with the person bounding box

When a person is detected, the coordinates for each bounding box of class person is published from YOLOv3 to the detected_objects_in_image topic [1]. From there, the custom program arm_commad.py is used to calculate the coordinates of the center of the bounding box, which is sent to ik_move.py to generate actuator commands for the manipulator.

The manipulator consists of three ROBOTIS Dynamixel XM430-W350 actuators, which were selected as they are one of the most commonly used ROS-compatible actuators. One actuator is used for yaw and two are used for pitch, forming a manipulator with 3 degrees of freedom (DOF). Although a ROS-compatible motion planning framework called MoveIt does exist, the program is optimized for arms with 6 or more DOFs. A custom inverse kinematics algorithm was developed because this is simpler and more reliable for a 3 DOF manipulator.

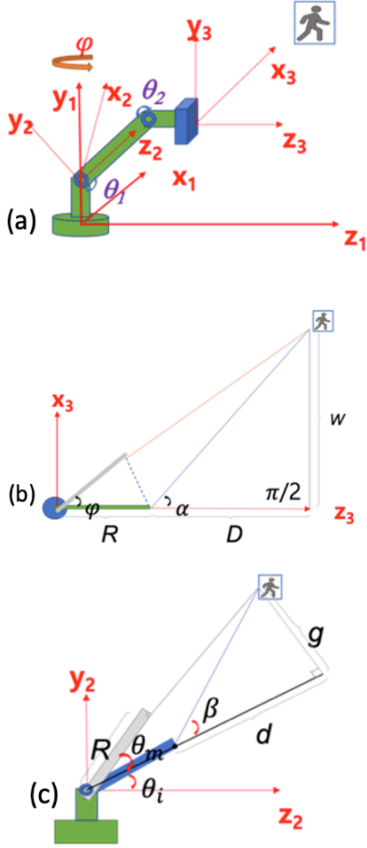

Fig. 7. (a) Coordinate axes of the manipulator, (b) top-down view of Fig. 7a, (c) side view of Fig. 7a

The algorithm works by aligning the center of the USB camera to the center of the bounding box in YOLO. As shown in Fig. 6, the point O has to align with point P. Moving the manipulator will shift O, while P will shift to follow the person.

The coordinate axes of the manipulator can be visualized as shown in Fig. 7a.

Yaw can be calculated according to Fig. 7b. Since R = ~ 0.1m and D = 1 ~ 2m, $R + D$ can be approximated as D with over 90% accuracy. Thus, $\varphi$ equals $\alpha$. The camera's maximum view angle is $\pi/3$ = 60° horizontally and vertically, and each image is 640 x 480 pixels. Then, $P_{x_{max}} = 320$, $\alpha_{max} = \pi/6$. For some small angle $\alpha$, (1) holds. As a result, yaw can be calculated with (2).

$$\frac{\alpha}{\alpha_{max}} = \frac{P_x}{P_{x_{max}}} \tag{1}$$

$$\varphi \approx \alpha = \frac{\pi}{6*320} * P_x \tag{2}$$

Fig. 7c shows how pitch can be calculated. The manipulator has 3 DOF, however this can effectively be simplified to a pan-tilt system since there is just one overall angle for the end effector to align with the person's forehead. Like above, d is significantly larger than R, so $R + d$ can be approximated as d. Then, $\tan(\theta_m) = \frac{g}{R+d} \approx \frac{g}{d} = \tan(\beta)$. For a small angle $\beta$, (3) Therefore, pitch can be calculated as (4).

$$\frac{\beta}{\beta_{max}} = \frac{P_y}{P_{y_{max}}} \tag{3}$$

$$\theta_m \approx \beta = \frac{\pi}{6*240} * P_y \tag{4}$$

### 2.3 Mobile Robot

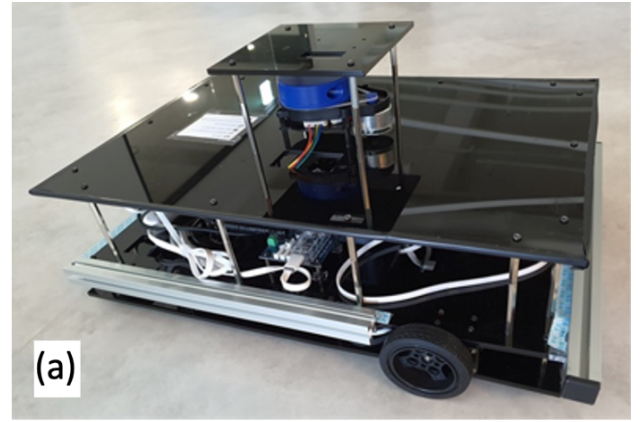

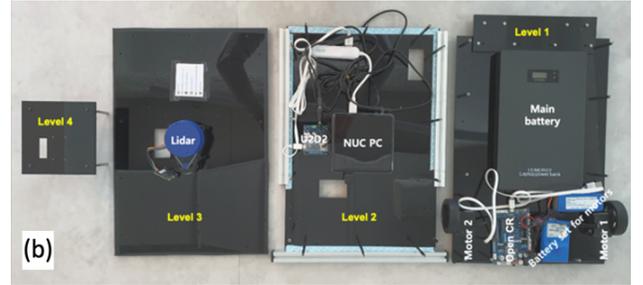

Fig. 8. (a) Fully assembled mobile robot sub-module, (b) Disassembled mobile robot sub-module.

The mobile robot provides a stable base for the manipulator and performs Simultaneous Localization and Mapping (SLAM), autonomous navigation, and path planning. SLAM is the problem where a mobile robot needs to both acquire a map (mapping) and know its location within that map (localization) at the same time [7]. Navigation is the problem of getting from one place to another and is dependent upon localization, path planning, and mapping [4].

The physical sub-module consists of the following components:
- Intel Nuc PC: Runs Ubuntu 18.04 and ROS 1 Melodic Morenia and was chosen for its high performance yet small size.
- x2 ROBOTIS Dynamixel XM430-W210: The actuators were chosen because of their ROS compatibility and for being one of the most commonly used actuators in ROS.
- OpenCR Board: Used for motor control and communication between Intel Nuc and Dynamixel XM430-W210s.
- U2D2 Board: Used for communication between manipulator's actuators and Intel Nuc.
- YDLIDAR X4 Lidar: Selected for its high quality, yet affordable price and ROS

compatibility. This is used for remote sensing, which is critical for SLAM, path planning, and navigation.
- iEnergy 2N Battery: Selected for large capacity, providing the necessary current and voltage, and for being able to simultaneously power the Nuc PC and manipulator without encountering overcurrent issues.

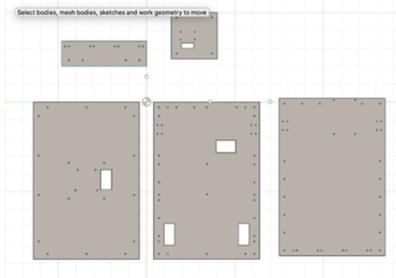

Fig. 9. CAD file of the plates for the mobile robot in Autodesk Fusion360

To carry all of the aforementioned components, Autodesk Fusion360 was used to design the frame of the custom mobile robot, as shown in Fig. 9a. 5mm thick black acrylic boards were chosen so that they could support the weight of the components without bending and the boards were cut to match the CAD schematics using the cutter in 9b. Two small omnidirectional ball casters were added and 20mm x 20mm aluminum profiles were added as guards.

A custom ROS package was also developed for the mobile robot. A new Uniform Robot Description Format (URDF) was devised to provide dimensions of the robot to ROS, which is necessary for visualization and path planning. The Gmapping algorithm [6] was used for lidar-based SLAM and the ROS navigation stack was incorporated for autonomous navigation. The custom program move_robot.py subscribes to the *person_present* topic and if an integer greater than 0 is received, the current goal is saved in save(), an instance of the move_base package's MoveBaseGoal [6]. Then, the action_lib package's cancel_goal() function is called and the mobile base stops [6]. When *person_present* returns 0, indicating that no person is detected, the callback function initiates the actionlib package's send_goal() function with the goal being the aforementioned save().

## 3. TESTING, CHALLENGES, AND SOLUTIONS

First, each component: end effector, manipulator, and mobile robot were tested separately. Then, the end effector and manipulator were connected, checked for compatibility, then attached to the mobile robot.

### 3.1 Testing End Effector

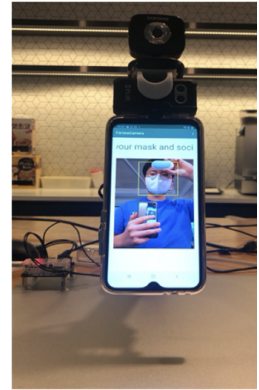

Fig. 10. Testing end effector sub-module with custom Android app

Firstly, the sample app from the SDK was tested for compatibility between the Samsung Galaxy M20 and the FLIR ONE Pro thermal camera. Next, a custom app was developed with a camera view, marquee text, and TTS engine which was again tested for compatibility. After that, the remaining functions were added and tested. When testing the end effector, three problems arose, and the following solutions were implemented:
- High temperatures were detected even though the person had normal temperature. This occurred when there were sources of temperature greater than the body within the display. This was resolved by using a detection algorithm to take coordinate values of the face and taking temperature values of those coordinates.
- The face tracking algorithm was too slow. The initial face tracking algorithm that was used was Google's ML Kit. However, this was problematic because all images had to be sent to Google's servers, where the image processing would occur, and results would be sent back to the device. This led to latency issues that made it unusable with real-time footage. Instead, the older Google Mobile Vision API was used as it processed images quickly enough to work alongside the real-time footage.
- The camera sporadically returned unreasonably high temperatures. To solve this issue the program was altered to check to see that if three consecutive temperature readings were above the threshold, only then would the chatbot activate.

### 3.2 Testing End Effector and Manipulator

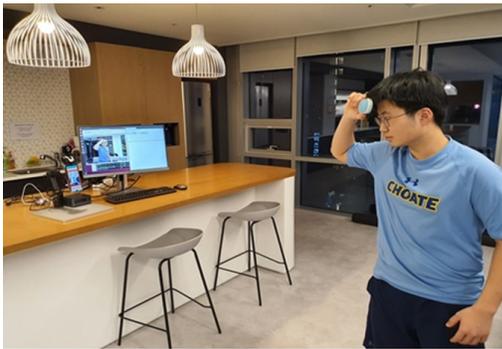

Fig. 11. Testing end effector and manipulator with a hot pack simulating a fever

Once the end effector sub-module was tested and empirically proven to work, it was combined with the manipulator and tested for compatibility. The objective was to test that the arm would correctly detect and follow people, while the end effector would successfully extract temperature values and engage in conversations. Originally, the ROS middleware for the manipulator and Android system for the end effector functioned separately. In this case, a successful outcome would only occur if all three end criteria were met:
- Condition A: YOLOv3 algorithm (manipulator) detected a person,
- Condition B: Google Mobile Vision API (end effector) detected a face,
- Condition C: FLIR thermal camera provided a temperature reading.

However, if the manipulator vibrated too much or if a person passed by too quickly, it was unable to detect the person's temperature. Therefore, to increase the likelihood of a successful reading, nine points were taken from each detection and the maximum temperature was calculated among the nine points. Furthermore, the conditions were simplified so that only one detection algorithm would be needed. The YOLOv3 algorithm running on the Nuc PC could be enhanced by adjusting weights, reducing the data set, and being retrained while modifying the Mobile Vision API was limited. Furthermore, the Mobile Vision API was less accurate and took longer to detect objects. So, the Mobile Vision API (Condition B) was removed in favor of the YOLOv3 algorithm. However, the removal of the Mobile Vision API meant that the YOLOv3 algorithm had to directly communicate with the Android phone to provide coordinates for the bounding box. This was accomplished by publishing coordinates to a web server and website, from which the Android app would use a web crawler to extract coordinate data, which would be used to calculate the temperature [2].

### 3.3 Testing Mobile Robot

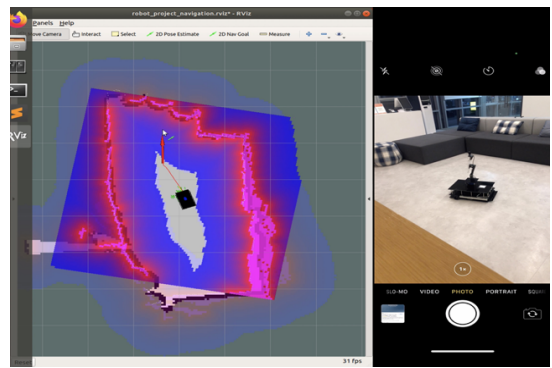

Fig. 12. Mobile robot performing autonomous navigation with Rviz visualization software on the left

When navigating autonomously, at times the mobile robot encountered localization issues, causing it to stop. The following adjustments were made to improve its performance [8]:
- Reduction of the robot's acceleration and top speed.
- Overall path split into smaller component paths
- Adjustments to Global Planner's *cost_factor* and *neutral_cost* to change *overall_cost*
- Dynamic Window Approach (DWA) Local Planner modified with changes to *sim_time*

### 3.4 Testing Full Robotic System

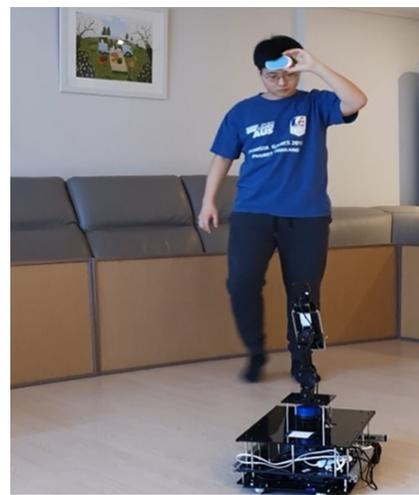

Fig. 13. Testing the full robotic system

After individually testing each sub-module, the end effector, manipulator, and mobile robot were connected and tested as a whole. During testing, two key problems were identified and resolved as follows:
- Manipulator's vibration made detection unreliable. When the robot was navigating autonomously, the manipulator vibrated significantly, which made it challenging to detect people and get a temperature reading. This issue was resolved by programming the mobile robot to temporarily stop when a person is detected. This provided the manipulator with

- a stable platform to scan, follow, and get people's temperature.
- Lidar detection of manipulator as an obstacle. When performing navigation, the lidar sensor often misinterpreted the robot arm as an obstacle, causing the mobile robot to freeze in place as it tried to recalculate a path. The lidar sensor has a minimum detection range of 120mm, so the robot arm was added to a fourth level and the supports for the level were placed within 120mm, effectively making them invisible to the lidar.

## 4. CONCLUSION

Considering the prevalence of increasingly contagious and deadly variants of COVID-19, and the slower than expected rates of vaccination, there is still a great need to be cautious and intensify all protective measures against the virus. With that in mind, there is a compelling need for a self-driving, temperature scanning, conversing mobile robot which can keep us all socially distanced, scanned, and safe. Throughout this research project, a mobile robotic solution has been presented that can navigate autonomously, detect and follow people, scan their temperature, and converse with them about potential symptoms of COVID-19. This research project acts as a feasibility study that demonstrates a functioning prototype, highlights challenges encountered with explanations on how they were resolved and provides suggestions on improvements required for commercialization.

## 5. FUTURE RESEARCH AREAS

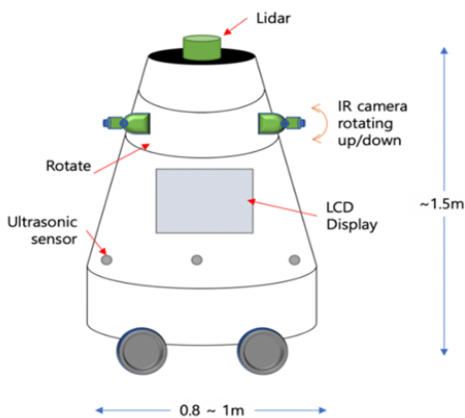

Fig. 14. Proposed design for an improved autonomous system that could be commercialized

The purpose of this project was to develop a robotic system that can navigate autonomously, find and follow human faces, extract temperature readings, and converse with users about COVID-19 symptoms. While the above project has successfully demonstrated this goal, enhancements could be made to the following aspects of the system:

- Robot Dimensions: The current robot has dimensions of 350mm x 500mm, with a maximum speed of 986m/h, significantly slower than the speed of a walking person. An increase in the size of the robot to roughly 800mm x 900mm with an increased maximum translational speed of around 5km/h would allow for greater stability and practicality in the real world.
- Remote Sensing: While the current lidar, the YDLIDAR X4, has decent resolution, it frequently detects noise and nonexistent obstacles. This leads to frequent pauses as the mobile robot attempts to recalculate a plan towards the desired goal and on occasion, the robot stops entirely. To resolve this issue, a lidar with superior noise filtering capabilities could be used or other sensors could be used in conjunction with the lidar for greater noise filtering leading to better navigation.
- Computer Vision: Using the YOLOv3 algorithm, the current iteration of the manipulator is capable of distinguishing people from other objects and is able to follow and get their temperature readings. However, if multiple people moving in different directions are detected simultaneously, the robot has difficulty deciding whom to follow first. Algorithmic improvements can be made so that not only is the robot able to decide in which order to scan people's temperatures but could also be linked to a database in a way that it is able to recognize whether a person has already been screened or not.
- Improved HCI: The current chatbot is only capable of brief conversations on vaccination and COVID-19 symptoms. The chatbot could be trained and improved to serve as a general conversational COVID-19 chatbot that can provide keep users updated with critical information and provide targeted recommendations.
- Incorporation of swarm techniques: The current system operates alone, however if swarm techniques were involved, this would allow for multiple such robots to simultaneously scan and communicate with each other. Especially when connected to a database of employee's faces, this would allow for more efficient and effective screening.

# APPENDIX

Table 1. Hardware Specifications

| Component | Specifications |
|---|---|
| Intel Nuc PC (10 Performance Kit \| NUC10i7FNH) | **Board number:** NUC10i7FNB, **DC Input Voltage:** 19VDC, **Max memory size:** 64GB, **Max memory bandwidth:** 42.6GB/s, **Dimensions:** 117mm x 112mm x 51mm |
| OpenCR Board | **Microcontroller:** STM32F746ZGT6 / 32-bit ARM Cortex®-M7 with FPU (216MHz, 462DMIPS), **Sensors:** Gyroscope 3Axis, Accelerometer 3Axis, Magnetometer 3Axis (MPU9250), **External input voltage:** 5 V (USB VBUS), 7-24 V (Battery or SMPS), **External Output voltage:** 12V@1A(SMW250-02), 5V@4A(5267-02A), 3.3V@800mA(20010WS-02), **Dimensions (WxD):** 105mm x 75mm, **Mass:** 60g |
| iEnergy 2N Battery | **Capacity:** 20,800mAh, **Size (LxWxH):** 205mm x 95mm x 25mm, **Mass:** 760g, **Input voltage:** DC 12V ~ 24V, **Output voltage:** 5V, 9V, 12V, 16V, 19V, **Charging time:** 4.5 hours |
| iEnergy S Battery | **Capacity:** 10,000mAh, **Dimensions (LxWxH):** 120mm x 80mm x 23mm, **Mass:** 304g, **Output voltage:** 5V, 9V, 12V |
| LIPO 11.1V, 1800mAh LB-012 Battery | **Capacity:** 1800mAh, **Dimensions:** 88mm x 35mm x 26mm, **Weight:** 106g, **Voltage:** 11.1V |
| Dynamixel XM430-W210 (for Mobile Base) | **MCU:** ARM CORTEX-M3 (72MHz, 32Bit), **Position Sensor:** Contactless absolute encoder (12Bit, 360°), **Motor:** Coreless, **Baud rate:** 9600 bps ~ 4.5Mbps, **Control algorithm:** PID Control, **Resolution:** 4096 pulses/rev, **Backlash:** 15 arcmin, **Weight:** 82g, **Dimensions (WxHxD):** 28.5mm x 46.5 mm x 34mm, **Gear Ratio:** 212.6 : 1, **Input Voltage:** 10.0 ~ 14.8V (12.0V recommended) |
| Dynamixel XM430-W350 (for Manipulator) | **MCU:** ARM CORTEX-M3 (72MHz, 32Bit), **Position Sensor:** Contactless absolute encoder (12Bit, 360°), **Motor:** Coreless, **Baud Rate:** 9600 bps ~ 4.5Mbps, **Control Algorithm:** PID Control, **Resolution:** 4096 pulses/rev, **Backlash:** 15 arcmin, **Weight:** 82g, **Dimensions (WxHxD):** 28.5mm x 46.5 mm x 34mm, **Gear Ratio:** 353.5 : 1, **Input Voltage:** 10.0 ~ 14.8V (12.0V recommended) |
| YDLIDAR X4 Lidar | **Range Frequency:** 5000Hz, **Scanning Frequency:** 6-12 Hz, **Range:** 0.12-10m, **Scanning Angle:** 0-360°, **Angle Resolution:** 0.48-0.52°, **Supply Voltage:** 4.8-5.2V, **Voltage Ripple:** 0-100 mV, **Starting Current:** 400-480mA, **Sleep Current:** 280-340mA, **Working Current:** 330-380mA, **Baud Rate:** 128000 bps, **Signal High:** 1.8-3.5V, **Signal Low:** 0-0.5 V, **PWM Frequency:** 10kHz, **Duty Cycle Range:** 50-100%, **Laser Wavelength** 775-795mm, **Laser Power:** 3 mW, **FDA:** Class I, **Working Temperature:** 0-40 °C, **Lightning Environment:** 0-2000 Lux, **Weight:** 189g, **Size:** 102mm x 71mm x 63mm |
| Samsung A-1200MB USB Camera | **Dimensions (LxWxH):** 48.2mm x 50.1mm x 61.4mm, **Mass:** 71g, **Frame rate:** 30FPS (max), **Image resolution:** 12,000,000 square pixels, **Video resolution:** 640 pixels x 380 pixels |
| Samsung Galaxy M20 | **Dimensions:** 156.4 mm x 74.5 mm x 8.8 mm, **Weight:** 186g, **Display Size:** 6.3 in, **Display Resolution:** 1080 x 2340 pixels, **Battery Capacity:** 5000mAh |
| FLIR ONE Pro Thermal Camera | **Accuracy:** ±3°C, **Dimensions (HxWxD):** 68mm x 34mm x 14mm, **Non-operating temperature:** -20°C ~ 60°C, **Scene Dynamic Range:** -20°C ~ 400°C, **Mass:** 36.5g, **Frame rate:** 8.7 Hz, **Thermal resolution:** 160 pixels x 120 pixels |
| Hand Warmer & Power Bank DM-500 | **Capacity:** 5200mAh, **Size (LxWxH):** 104mm x 53.5mm x 29mm, **Mass:** 132g, **Input Voltage:** 5V/2A |